\documentclass[letterpaper,10pt,conference]{IEEEtran}
\newif\ifanon

\usepackage{amsmath,amssymb,bm}
\usepackage{graphicx}
\usepackage{booktabs}
\usepackage{siunitx}
\usepackage[noadjust]{cite}
\usepackage{microtype}

\graphicspath{{figures/}}
\newcommand{\Ad}{\operatorname{Ad}}
\newcommand{\ad}{\operatorname{ad}}
\IEEEoverridecommandlockouts

\title{\LARGE\bf Cosserat Modeling of Trimmed Helicoid Soft Arms with a Separated-Section Constitutive Law}

\author{Zhihang Qin*, Linxin Hou, Zeyu Zhong, Yuchen Sun, Wenci Xin, Yueheng Zhang, Ji Qi \\  Jie Wang, Peiyi Wang, Muhammad Sunny Nazeer*, Yu Jun Tan*, Federico Renda*, and Cecilia Laschi*%
\thanks{This work was supported by the NUS bridging fund — AI-Driven Soft Robots for Marine and Unstructured Environments, WBS A-8003743-00-00 (Corresponding author: Zhihang Qin, Muhammad Sunny Nazeer, Yu Jun Tan, Federico Renda, Cecilia Laschi)}%
\thanks{Zhihang Qin, Zeyu Zhong, Yuchen Sun, Wenci Xin, Yueheng Zhang, Ji Qi, Peiyi Wang, Muhammad Sunny Nazeer, Yu Jun Tan, and Cecilia Laschi are with the Department of Mechanical Engineering, College of Design and Engineering, National University of Singapore, Singapore.}%
\thanks{Zhihang Qin, Linxin Hou, Zeyu Zhong, Yuchen Sun, Yu Jun Tan, and Cecilia Laschi are with the Advanced Robotics Centre, National University of Singapore, Singapore.}
\thanks{Linxin Hou is with the Department of Electrical and Computer Engineering, National University of Singapore, Singapore.}
\thanks{Jie Wang is with the Department of Mechanical Engineering, KU Leuven, Leuven, Belgium.}
\thanks{Federico Renda is with the Department of Mechanical and Nuclear Engineering, Khalifa University, Abu Dhabi, United Arab Emirates.}
}

\makeatletter
\g@addto@macro\normalsize{%
  \setlength{\abovedisplayskip}{4pt}\setlength{\belowdisplayskip}{4pt}%
  \setlength{\abovedisplayshortskip}{2pt}\setlength{\belowdisplayshortskip}{2pt}}
\makeatother

\begin{document}
\maketitle
\bstctlcite{IEEEexample:BSTcontrol}

\begin{abstract}
Cosserat rod models for soft robots usually construct sectional stiffness by summing material properties over a common cross-section. This assumption becomes inaccurate for trimmed helicoid arms, where load-bearing helix domains are separated and connected only through sparse fused crossings. This paper formulates a separated-section constitutive law that evaluates each helix domain in its local frame and pulls its constitutive response back to the backbone, yielding an effective backbone stiffness. Sparse-fusion mechanics captures the additional compliance caused by relative motion between neighboring domains and determines channel-wise reduction profiles $\eta_c(s/L)$ for bending, torsion, and extension. The resulting effective sectional stiffness is strongly anisotropic: bending and extension are reduced by about one order of magnitude, whereas torsion remains close to the effective backbone stiffness. The resulting sectional law is embedded in a geometrically exact dynamic Cosserat model with GVS discretization and routed-tendon actuation. Across 103 measured configurations, the three datasets give pooled normalized position errors of \SI{7.7}{\percent}, \SI{6.7}{\percent}, and \SI{7.8}{\percent}, while each full-arm solve requires approximately \SI{0.3}{s} on one CPU core (Intel Xeon, Cascade Lake, \SI{2.8}{GHz}), enabling rapid model-based planning, state and load estimation, and morphology--control co-design for architected soft robots.
\end{abstract}


\begin{figure*}[!t]
\centering
\includegraphics[width=\textwidth]{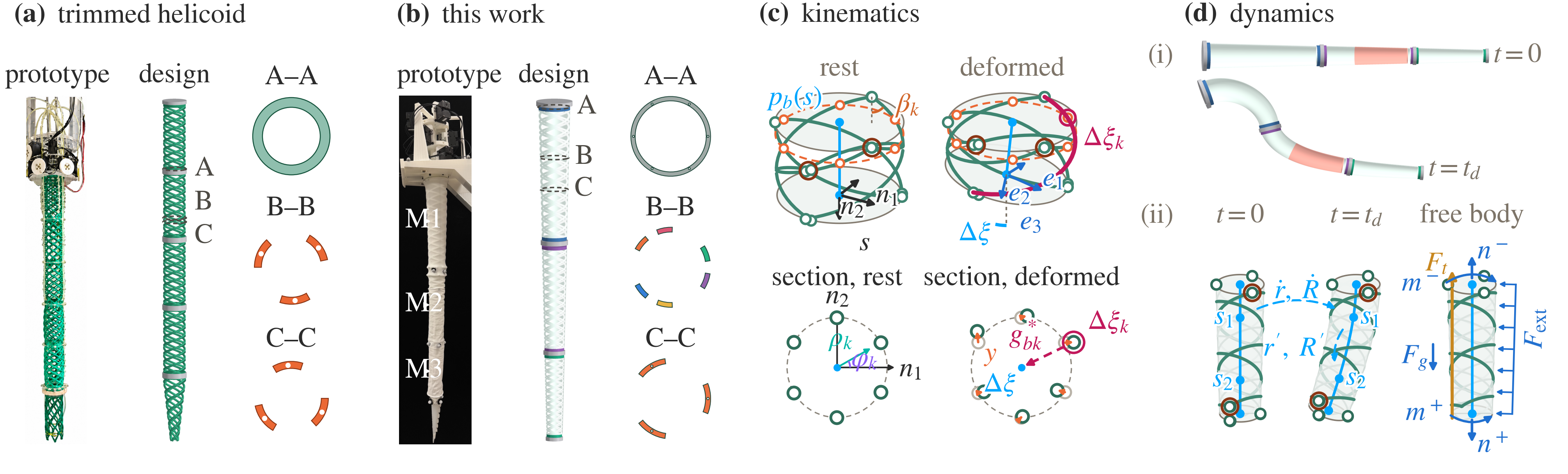}
  \caption{Trimmed-helicoid arms and their sectional topology.
    (a)~Trimmed-helicoid arm of \cite{guan2023trimmed}.
    (b)~This work: a three-arm-section tapered tendon-driven prototype
    (c)~Kinematics of one free arm segment: the domains move relative to the
    section frame, and the strain $\Delta\xi_k$ is pulled back to the
    backbone strain $\Delta\xi$.
    (d)~Dynamics: (i) the arm at rest and deformed, with the modelled segment
    shaded; (ii) the segment at $t=0$ and $t=t_d$, and as a free body with
    internal force $n^{\mp}$, internal moment $m^{\mp}$, external loads $F_{\mathrm{ext}}$, gravational force $F_g$ and tendon force $F_t$.}
\label{fig:platform}
\end{figure*}

\section{Introduction}

Soft and continuum manipulators exploit distributed body deformation to achieve large motion and compliant interaction. Octopus-inspired soft arms were an early example of this design principle and motivated both new robot architectures and continuum models \cite{laschi2012octopus,renda2012steady}. Their modeling later progressed from steady-state descriptions to full dynamics and model-based control \cite{renda2014dynamic,thuruthel2017dynamic}. More broadly, soft robotics has increasingly relied on physics-based reduced-order models to connect morphology, actuation, and control \cite{laschi2016technologies,mengaldo2022modeling}.

Cosserat rod theory provides a general one-dimensional framework for this purpose and has been widely used for tendon-driven continuum robots, multi-section soft manipulators, real-time dynamics, and geometric variable-strain simulation \cite{rucker2011statics,renda2018discrete,boyer2021dynamics,till2019real,mathew2022sorosim}. In these models, the sectional constitutive law maps local strain to internal force and moment, so the accuracy of the predicted deformation depends directly on how the sectional stiffness $K(s)$ represents the underlying load-bearing structure.

\begin{figure*}[!t]
\centering
\includegraphics[width=\textwidth]{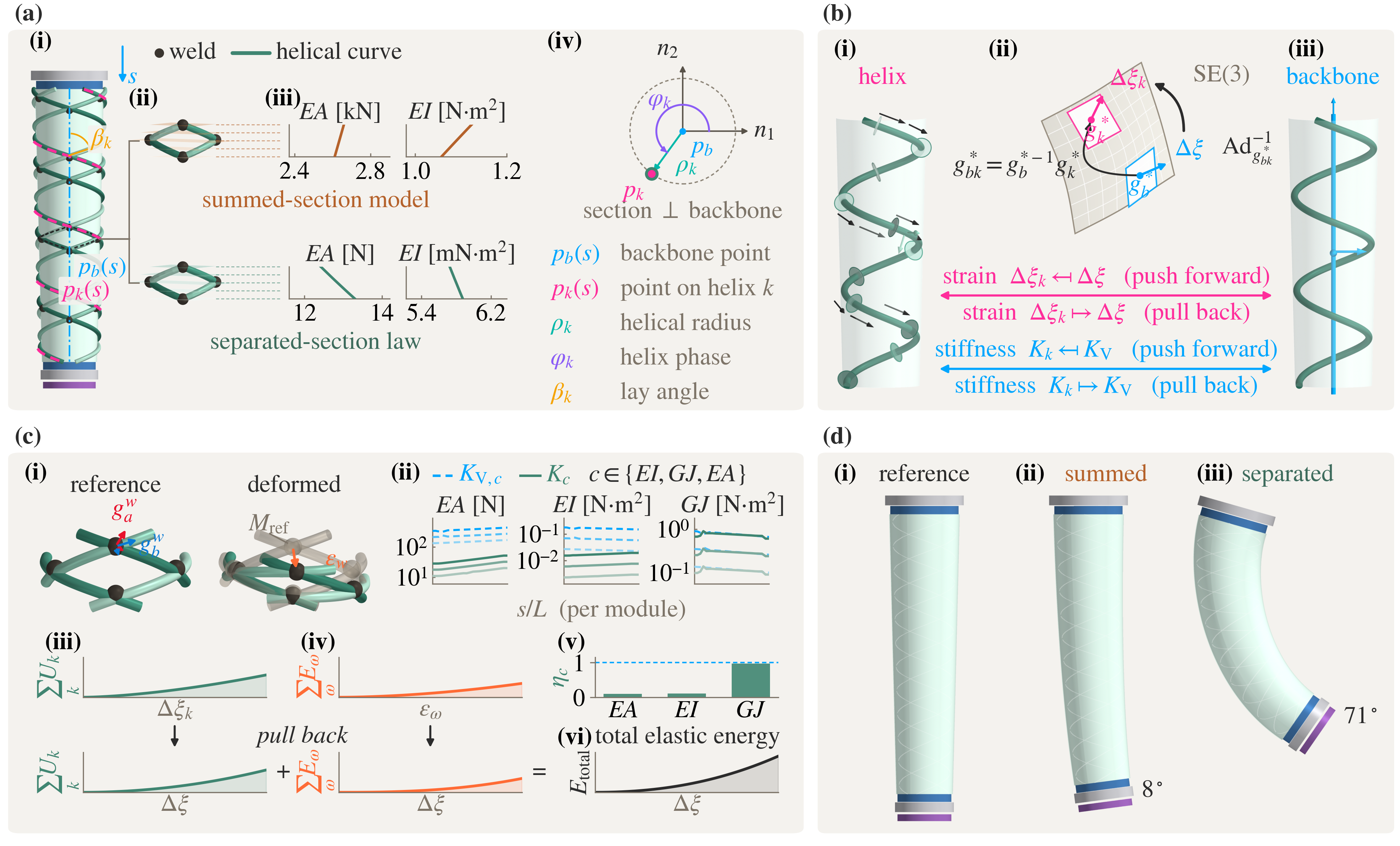}
\caption{Separated-section constitutive law. (a) Lattice geometry and sectional stiffness comparison. (b) Push-forward and pull-back mappings of strain and stiffness between helix and backbone frames. (c) (i) Reference and deformed fused crossings; (ii) stiffness distributions along each module: $K_{\mathrm{V},c}$ (dashed) denotes backbone stiffness under common sectional motion, and $K_c$ (solid) denotes effective stiffness after relaxation of relative helix-domain motion; the latter is substantially lower in bending and extension but nearly unchanged in torsion; (iii) helix deformation energy; (iv) fusion energy; (v) stiffness ratios $\eta_c=K_c/K_{\mathrm{V},c}$: 0.10 for extension, 0.11 for bending, and 0.98 for torsion; (vi) total elastic energy expressed in backbone strain. (d) Under \SI{9}{N} single-tendon tension, predicted end rotations are $8^\circ$ (summed-section model) and $71^\circ$ (separated-section law).}
\label{fig:law}
\end{figure*}

Recent studies have extended Cosserat models to more complex sections. Spatially varying material properties and fiber reinforcement can shift the effective stiffness center or neutral axis \cite{esser2025encoding,hanza2024inhomogeneous}, while additional sectional variables can represent cross-section inflation and contraction \cite{sun2024inflation}. These approaches still assume a common sectional motion. Trimmed helicoid arms introduce a different problem. Within each free segment, a backbone-normal cut intersects several separated load-bearing helix domains. These domains are connected only at sparse fused crossings, allowing neighboring domains to deform relative to one another [Fig.~\ref{fig:platform}(a,b)].

This topology appears in architectured soft manipulators, such as the trimmed-helicoid structure of Guan \emph{et al.} \cite{guan2023trimmed}. Patterson \emph{et al.} developed design-oriented stiffness models for trimmed helicoids, with an analytical approximation for compression and a semi-empirical model for bending \cite{patterson2025trimmed}. These models estimate the global stiffness of a helicoid module from geometric parameters, but do not provide a separated-section constitutive treatment for trimmed helicoid structures or integrate their mechanics into a Cosserat rod framework.

Structural mechanics provides the solutions needed to address these effects. Helical-structure models describe deformation and stiffness along helical load paths \cite{costello1997theory,cardou1997mechanical}, while shear-lag theory captures compliance associated with relative motion between connected components \cite{reissner1946shear}. Sectional and rod-lattice homogenization provide reduced descriptions of heterogeneous or discrete structures \cite{yu2002vabs,herrnbock2022homogenization,vinayak2026rod}. To the best of our knowledge, these ideas have not been combined to pull back the constitutive response of separated helix domains to a common backbone and integrate the resulting sectional law into a Cosserat rod model for trimmed helicoid soft robots.

Therefore, this paper addresses this problem in two steps. Firstly, each helix domain is evaluated in its local frame and its constitutive response is pulled back to the backbone, yielding the effective backbone stiffness $K_{\mathrm V}(s)$. Secondly, sparse-fusion mechanics accounts for the relative deformation permitted between fused crossings and determines channel-wise reduction factors $\eta_c(s/L)$ for bending, torsion, and extension. These factors reduce $K_{\mathrm V}$ to the effective sectional stiffness, which forms the sectional law embedded in a dynamic GVS Cosserat model with routed-tendon actuation.

The main contributions of this work are:

1) A separated-section constitutive formulation for trimmed helicoid soft arms that evaluates each helix domain in its local frame and pulls its constitutive response back to the backbone.

2) Integration of sparse-fusion mechanics to capture additional compliance caused by relative motion between neighboring helix domains and determine the effective bending, torsional and axial stiffness.

3) Integration of the resulting sectional law into a dynamic GVS Cosserat model, with experimental validation over 103 measured configurations, comparison against the conventional summed-section baseline, and analysis of distributed mechanical response and computational performance.

\section{Modeling Framework}
\label{sec:modeling}

Figure~\ref{fig:platform}(b--d) summarizes the prototype and the kinematic and dynamic descriptions used in the model. The arm is represented as a serial assembly of three tapered arm sections, denoted M1, M2, and M3, separated by rigid platforms. Each arm section is actuated by three tendons, while the distal tendons are routed through guide bores in the proximal sections. The backbone geometry, helix layout, tendon routes, mass distribution, and material properties are obtained from the CAD model and the fabricated structure. In the following derivation, $s$ denotes the backbone arc coordinate, $t$ denotes time, $(\cdot)'$ and $\dot{(\cdot)}$ denote differentiation with respect to $s$ and $t$, respectively, and $(\cdot)^*$ denotes the printed reference configuration. The hat map, $\Ad$, $\ad$, and $\ad^*$ denote the standard $\mathrm{SE}(3)$ map, adjoint, Lie-algebra adjoint, and coadjoint operators.

The central task is to obtain the sectional stiffness $K(s)$ required by the Cosserat model from the separated helix domains of each module. The local constitutive responses are evaluated first in the corresponding helix frames and pulled back to the backbone to obtain an effective backbone stiffness. Sparse-fusion mechanics then accounts for the additional compliance caused by relative deformation between neighboring domains. The resulting sectional law is finally integrated into a dynamic GVS Cosserat model of the complete tendon-driven arm.

\subsection{Sectional Mechanics of the Trimmed Helices}
\label{sec:law}

\subsubsection{Helix-to-backbone sectional mapping}

As illustrated in Fig.~\ref{fig:platform}(c), let $g_b^{*}(s)=(R_b^{*},p_b^{*})$ denote the backbone reference frame, with transverse directors $n_1$ and $n_2$. As illustrated in Fig.~\ref{fig:law}(a), the centerline of helix domain $k$ is parameterized by
\begin{equation}
\begin{aligned}
p_k={}&p_b+\rho_k
\left(
\cos\varphi_k\,n_1+\sin\varphi_k\,n_2
\right),\\
\varphi_k={}&\varphi_k^0+h_k\psi(s),
\end{aligned}
\label{eq:helix}
\end{equation}
where $\rho_k$ is the radial distance from the backbone to helix domain $k$, $\varphi_k$ is its azimuthal phase in the $n_1$--$n_2$ plane measured from $n_1$, $\varphi_k^0$ is the initial azimuthal phase, $\psi(s)$ describes the phase progression along the backbone, and $h_k=\pm1$ specifies the winding direction. With $s_k$ denoting the arc length along helix domain $k$,
\begin{equation}
\tan\beta_k=
\sqrt{(\rho_k\psi')^2+\rho_k'^2},
\qquad
\frac{\mathrm ds_k}{\mathrm ds}
=
\frac{1}{\cos\beta_k},
\label{eq:lay}
\end{equation}
where $\beta_k$ is the angle between the local helix tangent and the backbone tangent.

Each helix domain is evaluated on its own normal section [Fig.~\ref{fig:law}(a)],
\begin{equation}
K_k=
\operatorname{diag}
\left(
EI_{k1},EI_{k2},GJ_k,
GA_k,GA_k,EA_k
\right),
\label{eq:Kk}
\end{equation}
where $E$ and $G$ are the Young's and shear moduli, $A_k$ is the area of the section normal to the local helix tangent, $I_{k1}$ and $I_{k2}$ are its principal moments of area, and $J_k$ is its torsional constant. These sectional properties are obtained from the corresponding CAD sections.

Figure~\ref{fig:law}(b) illustrates the mapping between the backbone and helix-domain reference frames. Let $g_k^{*}$ denote the local reference frame of helix domain $k$, whose third axis is aligned with the local helix tangent. The relative transform from the backbone reference frame to helix domain $k$ is $g_{bk}^{*}=g_b^{*-1}g_k^{*}$.

Let $\xi=(\kappa;\nu)$ denote the backbone space twist, with $\kappa$ collecting two curvatures and torsion, and $\nu$ two shears and axial stretch. For the backbone material frame $g(s)\in\mathrm{SE}(3)$, $\xi=(g^{-1}g')^\vee$, and its printed reference value is $\xi^{*}=(g^{*-1}g^{*'})^\vee$, where $g^{*}(s)$ denotes the corresponding material frame in the reference configuration. The strain is the deviation $\Delta\xi=\xi-\xi^{*}$ from the printed twist. The corresponding strain $\Delta\xi_k$ in helix domain $k$ is
\begin{equation}
\Delta\xi_k
=
\cos\beta_k\,
\Ad_{g_{bk}^{*}}^{-1}
\Delta\xi ,
\label{eq:transport}
\end{equation}
where $\cos\beta_k=\mathrm ds/\mathrm ds_k$ converts the backbone-based strain measure to the helix arc-length measure.

The elastic energy per unit backbone length is
\begin{equation}
U'_{\mathrm V}
=
\frac12\sum_k
\frac{\mathrm ds_k}{\mathrm ds}
\Delta\xi_k^{\mathsf T}K_k\Delta\xi_k
=
\frac12
\Delta\xi^{\mathsf T}K_{\mathrm V}\Delta\xi .
\label{eq:voigtenergy}
\end{equation}

The quantity $U_k(\Delta\xi,y)$ used below denotes the elastic energy of helix domain $k$ when relative helix-domain motion $y$ is allowed; under the common-motion condition $y=0$, $\sum_k U_k(\Delta\xi,0)=\int U'_{\mathrm V}\,\mathrm ds$.

Using Eqs.~\eqref{eq:lay} and \eqref{eq:transport}, the local helix responses are pulled back to the backbone through energy equivalence, giving the effective backbone stiffness [Fig.~\ref{fig:law}(b)]
\begin{equation}
K_{\mathrm V}(s)
=
\sum_k
\cos\beta_k\,
\Ad_{g_{bk}^{*}}^{-\mathsf T}
K_k
\Ad_{g_{bk}^{*}}^{-1}.
\label{eq:voigt}
\end{equation}

The matrix $K_{\mathrm V}$ is the effective sectional stiffness obtained by pulling the helix-domain responses back to the backbone under a common sectional motion. It contains no identified parameter and differs from the summed-section baseline, which directly sums material properties on a backbone-normal cut.

For example, its axial entry is
\begin{equation}
(K_{\mathrm V})_{66}
=
\sum_k
\left(
EA_k\cos^3\beta_k
+
GA_k\cos\beta_k\sin^2\beta_k
\right),
\label{eq:axialentry}
\end{equation}
where the first term represents axial deformation along the helix tangent and the second represents transverse shear induced by backbone extension.

\subsubsection{Sparse-fusion correction and effective sectional stiffness}

The common-motion assumption overly limits the free-span structure because
neighboring helix domains can deform relative to one another between fused
crossings. As illustrated in Fig.~\ref{fig:law}(c)(i), at fused crossing
$w$, let $g_a^w$ and $g_b^w$ denote the material frames attached to the two
fused helix domains, and let $M_{\mathrm{ref}}$ denote their relative
transform in the reference configuration. The generalized fusion deformation is
\begin{equation}
\varepsilon_w=
\left[
\log\left(
M_{\mathrm{ref}}^{-1}(g_a^w)^{-1}g_b^w
\right)
\right]^{\vee},
\label{eq:fusionstrain}
\end{equation}
and the corresponding fusion energy is
\begin{equation}
E_w=
\frac12\varepsilon_w^{\mathsf T}K_w\varepsilon_w ,
\label{eq:fusionenergy}
\end{equation}
with
\begin{equation}
K_w=
\frac{\kappa_w}{\bar t}
\operatorname{diag}
\left(EI_w,EI_w,GJ_w,GA_w,GA_w,EA_w\right).
\label{eq:fusionstiffness}
\end{equation}
Here $K_w$ is the fusion stiffness, $\bar t$ is the characteristic fusion
length, and the subscript $w$ denotes the sectional properties of the fused
footprint. The scalar $\kappa_w$ corrects the nominal fusion stiffness and is
calibrated once from the tendon-free hanging configuration; it is then fixed
for all stiffness channels and subsequent simulations.

For arm section $\mu$, let $s_\mu^0$ denote its starting backbone coordinate
and $L_\mu$ its reference backbone length. The helix domains store the
elastic energies $\sum_k U_k$ [Fig.~\ref{fig:law}(c)(iii)], while the fused
crossings contribute $\sum_w E_w$ [Fig.~\ref{fig:law}(c)(iv)]. The total
elastic energy is
\begin{equation}
U_\mu=\sum_k U_k+\sum_w E_w .
\label{eq:totalenergy}
\end{equation}

Following classical beam sectional analysis \cite{giavotto1983anisotropic,yu2002vabs}, the internal sectional degrees of freedom are determined by minimizing the elastic energy subject to static equilibrium. Here, $y$ collects the relative-deformation coordinates of the separated helix domains. Taking the
reference configuration as stress-free and setting its elastic energy to
zero, the constant and first-order terms vanish. Under the assumed small
local material strains, retaining terms up to second order gives
\begin{equation}
U_\mu\simeq
\frac12
\begin{bmatrix}\Delta\xi\\y\end{bmatrix}^{\mathsf T}
\begin{bmatrix}
A & B^{\mathsf T}\\
B & D
\end{bmatrix}
\begin{bmatrix}\Delta\xi\\y\end{bmatrix},
\qquad
A=L_\mu K_{\mathrm V},
\label{eq:condensation_energy}
\end{equation}
where $B=\partial^2U_\mu/(\partial y\,\partial\Delta\xi)$ is the coupling
between the backbone strain and internal helix-domain motion. The internal
tangent stiffness contains both helix-domain and fusion contributions,
\begin{equation}
\begin{aligned}
D&=D_h+D_f,\\
D_h&=\frac{\partial^2}{\partial y^2}\left(\sum_kU_k\right),
\qquad
D_f=\sum_wG_w^{\mathsf T}K_wG_w ,
\end{aligned}
\label{eq:D}
\end{equation}
where $G_w=\partial\varepsilon_w/\partial y$ is the linearized mapping from
the internal coordinates to the fusion deformation.

For a prescribed backbone strain, internal equilibrium gives
\begin{equation}
\frac{\partial U_\mu}{\partial y}
=B\Delta\xi+Dy=0,
\qquad
y^*=-D^{-1}B\Delta\xi .
\label{eq:relaxation}
\end{equation}
Substitution of $y^*$ gives the relaxed total elastic energy
[Fig.~\ref{fig:law}(c)(vi)]
\begin{equation}
U_\mu^*=
\frac12\Delta\xi^{\mathsf T}
\left(A-B^{\mathsf T}D^{-1}B\right)\Delta\xi .
\label{eq:relaxedenergy}
\end{equation}
The corresponding sectional resultant is
\begin{equation}
\mathcal R=
\frac{1}{L_\mu}\frac{\partial U_\mu^*}{\partial\Delta\xi}
=
K^{\mathrm{lat}}\Delta\xi ,
\qquad
K^{\mathrm{lat}}
=
\frac{A-B^{\mathsf T}D^{-1}B}{L_\mu}
\preceq K_{\mathrm V}.
\label{eq:condensation}
\end{equation}

The bending, torsional, and axial channels of the directly calculated
effective backbone stiffness are
\begin{equation}
\begin{gathered}
K_{\mathrm V,EI}
=\tfrac12[(K_{\mathrm V})_{11}+(K_{\mathrm V})_{22}],\\
K_{\mathrm V,GJ}=(K_{\mathrm V})_{33},
\qquad
K_{\mathrm V,EA}=(K_{\mathrm V})_{66}.
\end{gathered}
\label{eq:channels}
\end{equation}
The corresponding condensed channels $K_c^{\mathrm{lat}}(s)$,
$c\in\{EI,GJ,EA\}$, are shown with $K_{\mathrm V,c}(s)$ in
Fig.~\ref{fig:law}(c)(ii) for $E=\SI{30}{\mega\pascal}$, $\nu=0.48$,
and $\kappa_w=4$. For each channel,
$\mathcal R_c=K_c^{\mathrm{lat}}\Delta\xi_c$. Under static equilibrium, $\mathcal R_c$ is equal
to the corresponding external force or moment resultant. Since
$K_{\mathrm V,c}(s)$ is obtained directly from Eq.~\eqref{eq:voigt}, a
measured load--strain pair directly determines the reduction. Defining
$\bar s_\mu=(s-s_\mu^0)/L_\mu$,
\begin{equation}
\eta_c(\bar s_\mu)
=
\frac{K_c^{\mathrm{lat}}(s)}{K_{\mathrm V,c}(s)}
=
\frac{\mathcal R_c(s)}
{K_{\mathrm V,c}(s)\Delta\xi_c(s)},
\qquad c\in\{EI,GJ,EA\},
\label{eq:eta}
\end{equation}
as illustrated in Fig.~\ref{fig:law}(c)(v). The effective sectional
stiffness used in the full-arm Cosserat model is
\begin{equation}
K_c^{lat}(s)=
\eta_c(\bar s_\mu)K_{\mathrm V,c}(s),
\qquad c\in\{EI,GJ,EA\}.
\label{eq:law}
\end{equation}

Although the arm undergoes large geometric deformation, the local material
response is assumed to remain in the small-strain regime. The extracted
off-diagonal sectional couplings are small and are therefore neglected, while
transverse shear is treated as rigid because the dominant compliance lies in
bending, torsion, and extension. With the same calibrated $\kappa_w$ used
for all three channels, bending and extension are reduced by approximately
one order of magnitude, whereas torsion remains close to the effective
backbone stiffness [Fig.~\ref{fig:law}(c)(ii),(v)].

\begin{figure*}[!t]
\centering
\includegraphics[width=\textwidth]{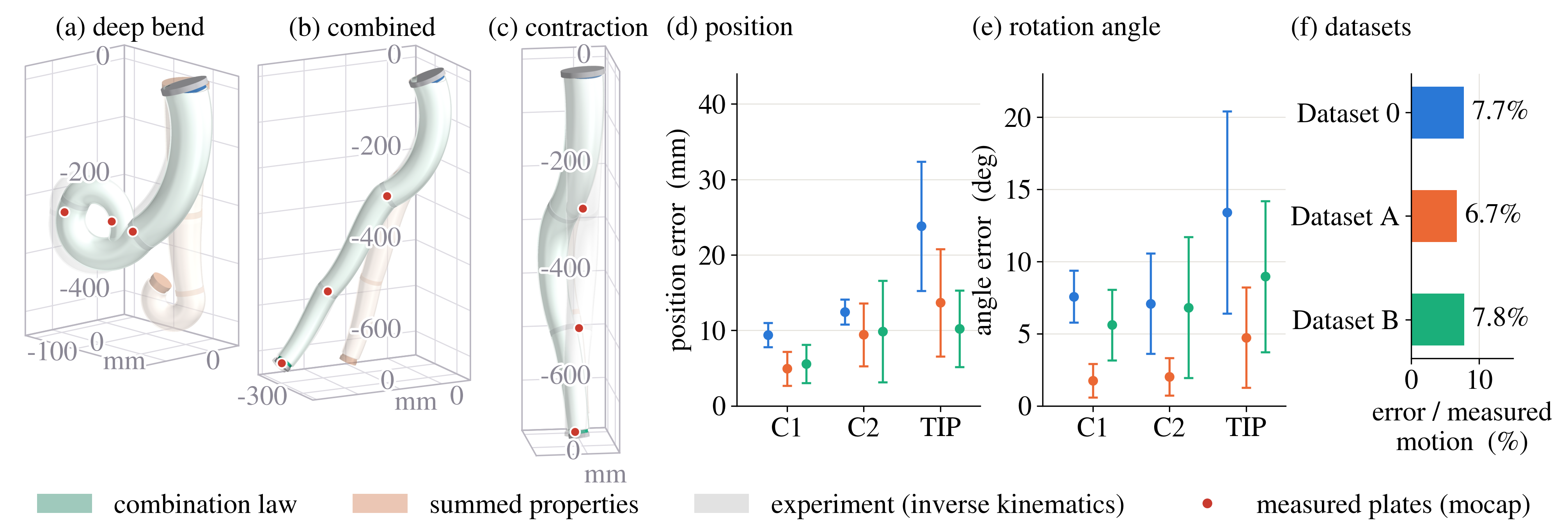}
\caption{Experimental validation over three deformation modes. (a--c) Deep bending, combined deformation, and axial contraction, with predictions from the effective sectional law and summed properties compared against experiment. (d,e) Position and rotation errors at C1, C2, and the tip. (f) Pooled normalized position errors of \SI{7.7}{\percent}, \SI{6.7}{\percent}, and \SI{7.8}{\percent} for Datasets 0, A, and B.}
\label{fig:experiment}
\end{figure*}

\subsection{Dynamic Cosserat Model and Numerical Method}
\label{sec:model}

\subsubsection{GVS dynamics}

Within each deformable section, $g(s,t)=(R,p)\in\mathrm{SE}(3)$ denotes the section pose, with orientation $R\in\mathrm{SO}(3)$ and backbone position $p\in\mathbb{R}^3$. The configuration satisfies
\begin{equation}
g'=g\widehat{\xi},
\qquad
\dot g=g\widehat{\eta},
\qquad
\eta'=\dot\xi-\operatorname{ad}_{\xi}\eta ,
\label{eq:kin}
\end{equation}
where $\eta=(\omega;v)$ is the body-velocity twist, with angular and linear components $\omega$ and $v$.

The GVS parametrization represents the strain field as
\begin{equation}
\xi
=
\xi^{*}
+
\Phi(s)q(t),
\qquad
\eta=J\dot q,
\qquad
J'=\Phi-\operatorname{ad}_{\xi}J ,
\label{eq:gvs}
\end{equation}

The internal wrench $\mathcal F=(m;n)$ combines the internal moment $m$ and force $n$:
\begin{equation}
\mathcal F
=
\mathcal R(s) + \Upsilon(s)\dot\xi
=
K(s)(\xi-\xi^{*})
+
\Upsilon(s)\dot\xi ,
\label{eq:internal}
\end{equation}
where $\Upsilon(s)$ is the sectional viscous matrix. The effective sectional stiffnesses $K_c^{lat}(s)$ obtained in Eq.~\eqref{eq:law} form the sectional stiffness matrix $K(s)$ used in the full-arm Cosserat dynamics.

Linear and angular momentum balance are combined in body coordinates as
\begin{equation}
\overline{\mathcal M}\dot\eta
+
\operatorname{ad}_{\eta}^{*}
\overline{\mathcal M}\eta
=
\mathcal F'
+
\operatorname{ad}_{\xi}^{*}\mathcal F
+
\mathcal F_e ,
\label{eq:strong}
\end{equation}
where $\overline{\mathcal M}(s)$ is the sectional inertia per unit length and $\mathcal F_e$ the distributed external wrench. Galerkin projection onto the GVS basis gives
\begin{equation}
M(q)\ddot q
+
[C(q,\dot q)+D_q]\dot q
+
K_q q
={}
F_g(q)+F_t(q,T)+
F_{\mathrm{ext}}(q),
\label{eq:reduced}
\end{equation}
where $M$, $C$, $D_q$, and $K_q$ are the generalized mass, velocity-dependent inertial, damping, and stiffness matrices, respectively; $F_g$, $F_t$, and $F_{\mathrm{ext}}$ are the gravity, tendon, and other external generalized loads, and $T$ collects the tendon tensions.

With $L$ denoting the total modeled backbone length, the principal reduced operators are

\begin{equation}
\begin{gathered}
M(q)=\int_0^L J^{\mathsf T}\overline{\mathcal M}J\,\mathrm ds
+\sum_\alpha J_\alpha^{\mathsf T}\overline{\mathcal M}_\alpha J_\alpha,\\
D_q=\int_0^L\Phi^{\mathsf T}\Upsilon\Phi\,\mathrm ds,\qquad
K_q=\int_0^L\Phi^{\mathsf T}K\Phi\,\mathrm ds .
\end{gathered}
\label{eq:operators}
\end{equation}

Here $\alpha$ indexes the rigid platforms, while $J_\alpha$ and $\overline{\mathcal M}_\alpha$ are their body Jacobians and lumped inertias. The matrix $C(q,\dot q)$ follows from the configuration-dependent mass matrix using the standard GVS formulation \cite{boyer2021dynamics}.

\subsubsection{Routed-tendon actuation}

Tendon $i$ follows the material offset $\rho_i(s)$, with spatial path and material-frame tangent
\begin{equation}
r_i=p+R\rho_i,\qquad
\tau_i=\nu+\kappa\times\rho_i+\rho_i',
\label{eq:route}
\end{equation}
and routed length and its gradient with respect to $q$,
\begin{equation}
\ell_i(q)=\int_0^L\|\tau_i\|\,\mathrm ds,\qquad
\nabla_q\ell_i=
\int_0^L\Phi^{\mathsf T}
\begin{pmatrix}
\rho_i\times\hat t_i\\
\hat t_i
\end{pmatrix}\mathrm ds ,
\label{eq:length}
\end{equation}
where $\hat t_i=\tau_i/\|\tau_i\|$ is the unit tendon tangent.

Channel friction gives the cumulative turning angle and tension field
\begin{equation}
\Theta_i(s)=\int_0^s\|\hat t_i'(\sigma)\|\,\mathrm d\sigma,
\qquad
T_i(s)=T_{i0}e^{-\mu_c\Theta_i(s)},
\label{eq:friction}
\end{equation}
where $T_{i0}=T_i(0)$ is the channel-entry tension and $\mu_c$ is the tendon--channel friction coefficient. The corresponding virtual work gives
\begin{equation}
F_t=
-\sum_i\int_0^L
\Phi^{\mathsf T}T_i(s)
\begin{pmatrix}
\rho_i\times\hat t_i\\
\hat t_i
\end{pmatrix}
\mathrm ds .
\label{eq:tendonforce}
\end{equation}

For commanded length $\ell_i^{\mathrm{cmd}}$, let
$e_i=\ell_i-\ell_i^{\mathrm{cmd}}$. The pull-only constraint is
\begin{equation}
e_i\le0,\qquad
T_{i0}\ge0,\qquad
T_{i0}e_i=0,
\label{eq:tendonkkt}
\end{equation}
and is enforced by
\begin{equation}
T_{i0}^{(r+1)}
=
\operatorname{clip}_{[0,T_{\max}]}
\left(\lambda_i^{(r)}+k_{\mathrm{AL}}e_i\right),
\qquad
\lambda_i^{(r+1)}=T_{i0}^{(r+1)},
\label{eq:alupdate}
\end{equation}
where $\lambda_i$, $r$, $k_{\mathrm{AL}}$, and $T_{\max}$ denote the multiplier, outer iteration, augmentation stiffness, and tension bound, while $\operatorname{clip}$ projects onto $[0,T_{\max}]$.

\subsubsection{Geometric saturation}
Large bending progressively closes the gaps between neighboring helix domains. For arm section $\mu$, the minimum clearance between monitored non-fused helix pairs is
\begin{equation}
\gamma_\mu(q)=\min_{(a,b)\in\mathcal P_\mu}\left(\|x_a-x_b\|-2r_s\right),
\label{eq:gap}
\end{equation}
where $x_a$ and $x_b$ are sampled helix-centerline positions, $r_s$ is the effective helix cross-sectional radius, and $\mathcal P_\mu$ is the monitored pair set. A compaction threshold is prescribed from the lattice geometry rather than fitted to the validation data. Once this threshold is reached, the generalized strain coordinates of that arm section are held fixed while the remaining coordinates continue to evolve. This geometric saturation constraint approximates section locking without an additional contact or jamming constitutive model.

\subsubsection{Discretization and numerical propagation}

The serial arm is represented by three deformable lattice modules and three \SI{8}{mm} transition sections adjacent to the rigid platforms; the arm section lengths are $273.5$, $216.0$, and $170.5$~mm. The modules use shifted Legendre strain bases of orders $(6,6,6,2,2,2)$ for $(\kappa_1,\kappa_2,\kappa_3,\nu_1,\nu_2,\nu_3)$, giving 30 generalized coordinates per module. Each transition uses a piecewise-constant six-component strain field, giving
\begin{equation}
n_q = 3\times30 + 3\times6 = 108 ,
\label{eq:nq}
\end{equation}
where $n_q$ is the total number of generalized coordinates.

Sectional integrals are evaluated by Gauss--Legendre quadrature, with intervals split at module boundaries, rigid connectors, and stiffness-profile breakpoints to resolve the spatial variation of $K(s)$ and the routed tendon geometry.

The pose and geometric Jacobian are propagated together using fourth-order Zanna--Magnus integration. For an interval of length $\Delta s$,
\begin{equation}
\Omega
=
\frac{\Delta s}{2}
(\xi_1+\xi_2)
+
\frac{\sqrt{3}\Delta s^2}{12}
\operatorname{ad}_{\xi_1}\xi_2 ,
\label{eq:zanna}
\end{equation}
where $\Omega$ is the Magnus twist increment and $\xi_1,\xi_2$ are evaluated at the two Zanna points. The propagation is
\begin{equation}
\begin{aligned}
g_{j+1}
&=
g_j\exp(\widehat{\Omega}),\\
J_{j+1}
&=
\operatorname{Ad}_{\exp(\widehat{\Omega})}^{-1}
J_j
+
\operatorname{dexp}_{-\Omega}\Phi_{\Omega},
\end{aligned}
\label{eq:propagation}
\end{equation}
where $j$ indexes the spatial integration intervals, $\Phi_{\Omega}$ is the corresponding basis increment, and $\operatorname{dexp}_{-\Omega}$ is the differential of the exponential map.

\section{Model Analysis and Experimental Validation}
\label{sec:validation}

\subsection{Experimental Setup}

The experimental platform consists of three tapered trimmed helicoid modules printed from TPU 95A using a Bambu Lab H2D printer, with rigid PLA connectors and mounting components. The arm is mounted vertically on a stainless-steel frame through a rigid bolted support. Each module is actuated by three tendons driven by ROBOTIS DYNAMIXEL XM430-W350-R servo motors through a ROBOTIS U2D2 interface, with tendon commands sent from Python using the \texttt{pySerial} library.

The deformable modules are enclosed by a thin flexible sleeve with negligible influence that follows the intended bending, twisting, and axial contraction. Arm motion is measured using an OptiTrack system running Motive: Tracker. Passive retroreflective markers on the two inter-module platforms and distal platform define the three tracked stations C1, C2, and tip, from which position and orientation are measured.

\begin{figure*}[!t]
\centering
\includegraphics[width=\textwidth]{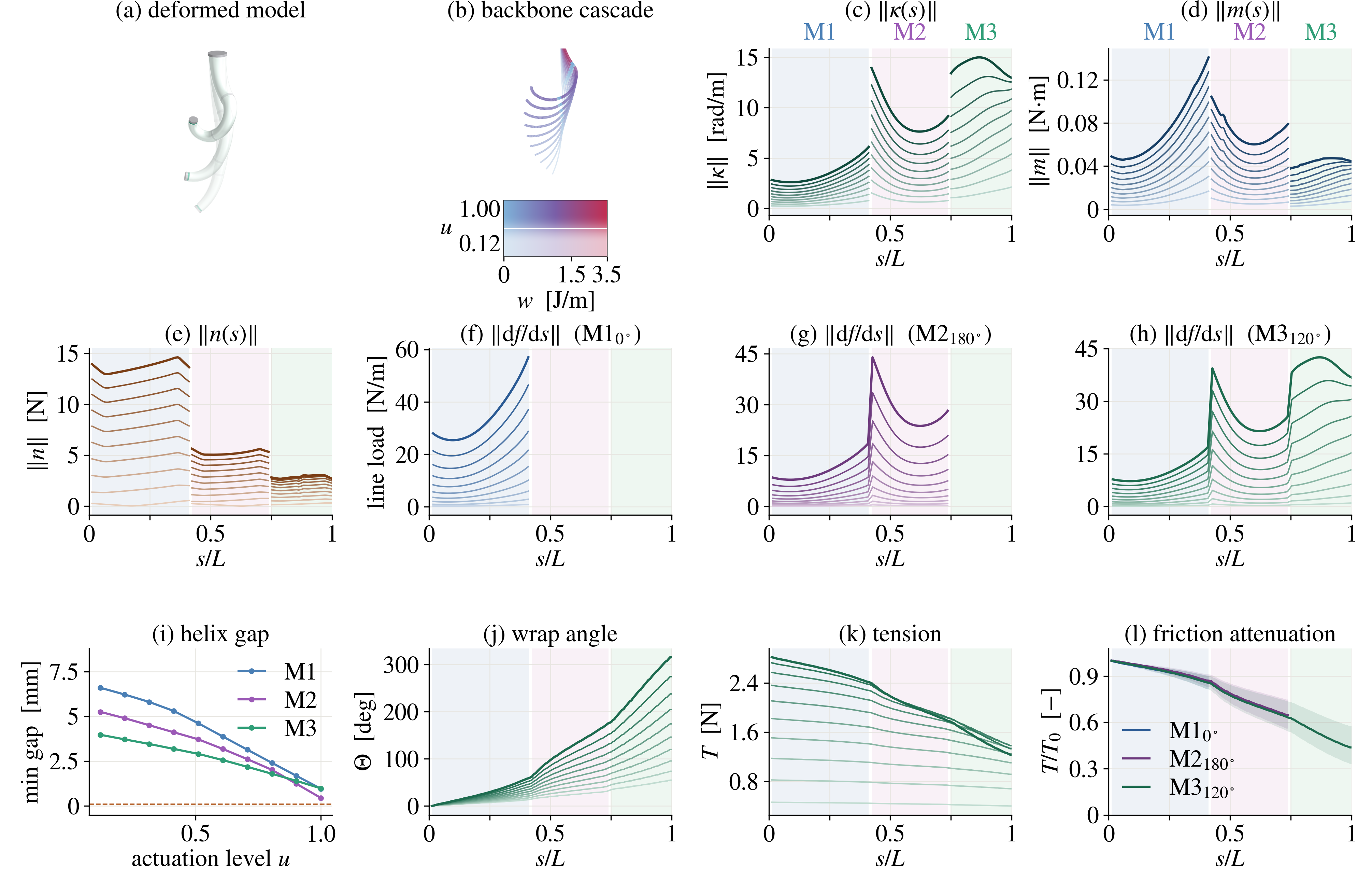}
\caption{Distributed response during three-tendon actuation. Tendons $\mathrm{M1}_{0^\circ}$, $\mathrm{M2}_{180^\circ}$, and $\mathrm{M3}_{120^\circ}$ retract by $u\times(140,185,270)\,\mathrm{mm}$, respectively, at ten levels spanning $u=0.12$--$1.00$. (a) Predicted configurations at $u=0.32$, $0.61$, and $1.00$. (b) Backbone evolution: hue indicates elastic energy per unit length $w(s)$ and lightness indicates $u$. (c--e) Rotational strain, internal moment, and internal force magnitudes. (f--h) Tendon line-load magnitudes $\|\mathrm{d}f/\mathrm{d}s\|$ along the three routes. Darker curves indicate increasing $u$ in (b--h,j,k); spatial plots shade module spans and omit interface segments. (i) Minimum non-fused helix gap in each module versus $u$; the dashed line marks the \SI{0.10}{mm} compaction threshold. (j--l) Frictional transmission post-processed from frictionless equilibria: (j) cumulative wrap angle $\Theta(s)$ of $\mathrm{M3}_{120^\circ}$; (k) its tension $T=T_0e^{-\mu_c\Theta(s)}$ at $\mu_c=0.15$, with entry tension $T_0$; (l) $T/T_0$ along all three routes at $u=1$, with solid curves for $\mu_c=0.15$ and bands for $\mu_c\in[0.10,0.20]$.}
\label{fig:fields}
\end{figure*}

\subsection{Validation Protocol and Parameter Identification}

The experiments comprise 103 configurations that span distal-tendon deep bending (Dataset 0), combined deformation (Dataset A), and axial contraction (Dataset B), yielding 309 platform-pose samples at C1, C2, and the tip. In the experiments, servo retractions follow successive linear ramps, each followed by a hold until the arm settles. All simulations reproduce these ramp durations and hold intervals, using measured tendon-length changes as inputs to both forward models. Platform poses recorded after settling are used only for evaluation; Table~\ref{tab:ablation} reports the normalized position errors at representative configurations.

For plate $b$ at configuration $k$, displacement is represented by its axial and lateral components to avoid dependence on bench--model azimuth alignment,
\begin{equation}
\bm\delta_{b,k}=
\left(
\Delta\bm p_{b,k}\!\cdot\!\hat{\bm u},
\left\|\Delta\bm p_{b,k}-
(\Delta\bm p_{b,k}\!\cdot\!\hat{\bm u})\hat{\bm u}\right\|
\right),
\label{eq:errorsplit}
\end{equation}
where $b\in\{\mathrm{C1},\mathrm{C2},\mathrm{TIP}\}$, $\Delta\bm p_{b,k}$ is the displacement from the upright reference, and $\hat{\bm u}$ is the arm axis. The position error is
\begin{equation}
\begin{aligned}
\mathrm{RMSE}_b&=
\sqrt{\frac{1}{2N_b}\sum_k
\|\bm\delta^{\mathrm{mod}}_{b,k}-\bm\delta^{\mathrm{meas}}_{b,k}\|^2},\\
\varepsilon_b&=
100\sqrt{\frac{\sum_k
\|\bm\delta^{\mathrm{mod}}_{b,k}-\bm\delta^{\mathrm{meas}}_{b,k}\|^2}
{\sum_k\|\bm\delta^{\mathrm{meas}}_{b,k}\|^2}},
\end{aligned}
\label{eq:positionerror}
\end{equation}
where $N_b$ is the number of configurations and $\mathrm{mod}$/$\mathrm{meas}$ denote prediction/measurement. The corresponding rotation error is
\begin{equation}
\mathrm{RMSE}^{\theta}_b=
\sqrt{\frac{1}{N_b}\sum_k
(\theta^{\mathrm{mod}}_{b,k}-\theta^{\mathrm{meas}}_{b,k})^2},
\qquad
\varepsilon_b^\theta=
100\frac{\mathrm{RMSE}^{\theta}_b}{\theta_b^{\max}},
\label{eq:rotationerror}
\end{equation}
where $\theta_{b,k}$ is the rotation from the upright reference and
$\theta_b^{\max}=82.6^\circ,100.2^\circ,111.7^\circ$ for C1, C2, and the tip, respectively, fixed across all datasets.

The three plate-wise position errors are pooled as
\begin{equation}
\varepsilon_{\mathrm{overall}}
=
100\sqrt{
\frac{\sum_b\sum_k
\|\bm\delta^{\mathrm{mod}}_{b,k}-\bm\delta^{\mathrm{meas}}_{b,k}\|^2}
{\sum_b\sum_k
\|\bm\delta^{\mathrm{meas}}_{b,k}\|^2}},
\label{eq:overallerror}
\end{equation}
giving the measured-motion-weighted dataset error in Fig.~\ref{fig:experiment}(f).

The fusion scale $\kappa_w$ is identified only once from the tendon-free hanging configuration. No tendon-driven configuration is used for sectional fitting; the resulting sectional stiffness is fixed throughout the validation.

\subsection{Full-Arm Prediction and Sectional-Stiffness Ablation}

Figure~\ref{fig:experiment}(a--c) compares the effective sectional law and summed properties under identical experimentally measured tendon inputs. The effective sectional law reproduces the deep coiling of Dataset 0, the combined multi-module deformation of Dataset A, and the axial contraction of Dataset B. In contrast, summed properties systematically overestimate the sectional stiffness and strongly underpredict all three deformation modes.

\begin{table}[!t]
\caption{Normalized position errors at representative configurations.}
\label{tab:ablation}
\centering
\scriptsize
\setlength{\tabcolsep}{3.0pt}
\begin{tabular}{@{}llccc@{}}
\toprule
Case & Sectional law & C1 & C2 & Tip \\
\midrule
Deep bend
& Effective sectional law & 9.7\% & 6.4\% & 7.9\% \\
& Summed properties & 91.5\% & 60.8\% & 25.3\% \\
Combined
& Effective sectional law & 2.8\% & 2.3\% & 1.5\% \\
& Summed properties & 83.5\% & 85.5\% & 85.7\% \\
Contraction
& Effective sectional law & 5.7\% & 4.3\% & 10.3\% \\
& Summed properties & 79.7\% & 78.4\% & 74.8\% \\
\bottomrule
\end{tabular}
\end{table}

Table~\ref{tab:ablation} quantifies the same difference in the representative states. Across the three tracked stations in the three representative cases, the effective sectional law gives normalized errors of \SIrange{1.5}{10.3}{\percent}, compared to \SIrange{25.3}{91.5}{\percent} for summed properties. The contrast is especially clear in Dataset A, where the summed-properties errors exceed \SI{83}{\percent} at all three plates.

Throughout the complete actuation sequences, Fig.~\ref{fig:experiment}(d,e) reports the station-wise position and rotation errors of the effective sectional law model, while Fig.~\ref{fig:experiment}(f) gives pooled position errors of \SI{7.7}{\percent}, \SI{6.7}{\percent}, and \SI{7.8}{\percent} for Datasets 0, A, and B. Since both models use the same measured tendon inputs and otherwise identical formulations, the ablation isolates the sectional closure and shows that the conventional summed-property assumption cannot recover the measured compliance.

\subsection{Distributed Mechanical Response}

Beyond the tracked poses, the model resolves how deformation and load are distributed along the arm. Figure~\ref{fig:fields}(a--e) shows that increasing tendon actuation redistributes curvature, strain energy, and internal wrench across the three modules, providing the mechanical state underlying the observed full-arm deformation.

\begin{figure}[!t]
\centering
\includegraphics[width=\columnwidth]{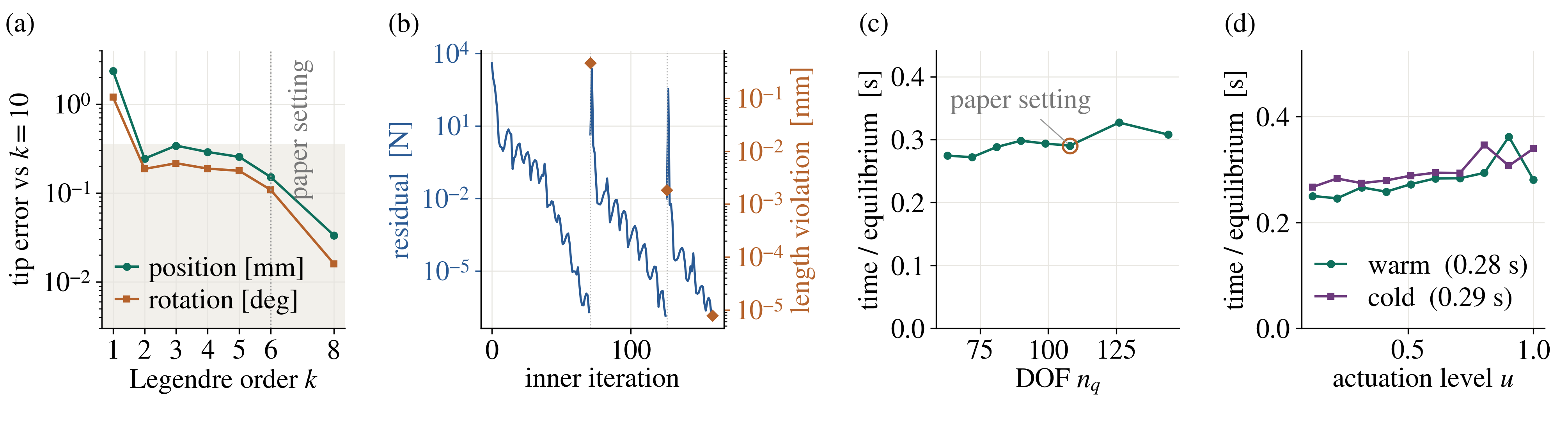}
\caption{Reduced-order solver convergence and cost. (a) Tip-position/rotation errors relative to $k=10$; shading marks the equilibrium-tolerance floor. (b) Deep-bending residual and tendon-length violation; dotted lines separate augmented-Lagrangian passes. (c) Median runtime versus coordinate count. Dotted line (a) and circle (c) mark $k=6$, $n_q=108$. (d) Warm/cold-start runtimes along Fig.~\ref{fig:fields}, with respective single-core medians near 0.3\,s.}
\label{fig:solver}
\end{figure}

This redistribution is coupled to the routed tendon geometry. The line loads in Fig.~\ref{fig:fields}(f--h) increase where tendon direction changes rapidly, particularly near highly curved regions and module transitions. Meanwhile, the minimum gap between non-fused domains decreases toward the prescribed threshold [Fig.~\ref{fig:fields}(i)], linking large bending to progressive lattice compaction.

The deformed tendon paths also determine the accumulated wrap angle and the associated frictional tension transmission. Figures~\ref{fig:fields}(j--l) show this effect for $\mu_c=0.15$, with $\mu_c\in[0.10,0.20]$ indicating its sensitivity to the channel-friction coefficient.

\subsection{Numerical Convergence and Computational Performance}

Figure~\ref{fig:solver} evaluates spatial resolution, tendon-constraint accuracy, and computational cost. Relative to a degree-ten reference, the tip discrepancy decreases from approximately \SI{2.4}{mm}/$1.2^\circ$ at degree one to \SI{0.03}{mm}/$0.016^\circ$ at degree eight [Fig.~\ref{fig:solver}(a)], placing the degree-six basis used throughout the simulations within the converged regime.

For the representative deep-bending case, the projected augmented-Lagrangian procedure reduces the inner residual by approximately ten orders of magnitude, while the tendon-length violation falls below \SI{1e-8}{m} within three outer updates [Fig.~\ref{fig:solver}(b)]. Therefore, the commanded tendon lengths remain accurately enforced during large deformation.

Computational cost changes only weakly with model resolution. All timings are measured on a single core of an Intel Xeon (Cascade Lake, \SI{2.8}{GHz}) processor. Increasing $n_q$ from 63 to 144 changes the median time from approximately \SI{0.27}{s} to \SI{0.31}{s}, while the $n_q=108$ model requires \SIrange{0.24}{0.37}{s} per commanded state. The solution times for both warm and cold starts remain close to \SI{0.3}{s} throughout the actuation sequence [Fig.~\ref{fig:solver}(c,d)], showing that deep bending and lattice compaction do not produce a corresponding increase in computational cost.

\section{Conclusion}

We introduced sectional mechanics for separated and obliquely oriented load-bearing domains into the Cosserat rod modeling of trimmed helicoid soft arms. Each helix is evaluated on its own normal section and transported into the backbone frame, while sparse fused crossings account for the additional compliance caused by relative domain motion. The resulting bending, torsional, and axial stiffnesses provide sectional closure for a dynamic GVS Cosserat model with routed-tendon actuation.

A single calibrated fusion correction yields a strongly anisotropic sectional prediction: bending and extension decrease to roughly one tenth of their rigid-transport values, whereas torsion remains much closer to the rigid-transport reference. Under the same measured tendon inputs, the effective sectional law reproduces deep bending, coupled deformation, and axial contraction, while the summed-property model systematically underpredicts the deformation. Across 103 measured configurations, the pooled normalized position errors remain below \SI{8}{\percent}. The model also resolves distributed curvature, internal wrench, tendon loading, and lattice compaction while requiring approximately \SI{0.3}{s} per simulated state on one CPU core.

This combination of structural fidelity and computational efficiency extends reduced-order Cosserat modeling toward more complex architected soft robots whose internal load paths cannot be represented by a rigid common section. It also creates a practical basis for rapid model-based trajectory planning and control, mechanics-aware state and load estimation, and morphology--control co-design over large configuration spaces.


\bibliographystyle{IEEEtran}
\bibliography{refs}

\end{document}